\documentclass[11pt,a4paper]{article}
\usepackage[hyperref]{acl2018}
\usepackage{times}
\usepackage{latexsym}

\usepackage{mathrsfs, amsmath, amssymb}
\usepackage{xspace}
\usepackage{url}
\usepackage{scrextend}
\usepackage{booktabs}
\usepackage{paralist}
\usepackage[tight-spacing=true]{siunitx}
\usepackage{color,soul}
\usepackage{etoolbox,multirow,adjustbox}
\usepackage{todonotes}
\robustify\bfseries

\aclfinalcopy % Uncomment this line for the final submission
\def\aclpaperid{1457} %  Enter the acl Paper ID here

\newcommand{\dataset}[1]{\textsf{#1}\xspace}
\newcommand{\resource}[2][]{\textsc{#2}#1\xspace}

\newcommand{\tabref}[1]{Table~\ref{#1}\xspace}
\newcommand{\figref}[1]{Figure~\ref{#1}\xspace}

\newcommand{\equref}[1]{Equation~\ref{#1}\xspace}

\newcommand\xent{\mathcal{X}}
\newcommand\hvec{\mathbf{h}}
\newcommand\yvec{\mathbf{y}}
\newcommand\xvec{\mathbf{x}}
\newcommand\D{\operatorname{D}}

\title{Towards Robust and Privacy-preserving Text Representations}

\author{Yitong Li \qquad Timothy Baldwin \qquad Trevor Cohn  \\
  School of Computing and Information Systems\\
  The University of Melbourne, Australia \\
 \tt yitongl4@student.unimelb.edu.au \\ \tt \{tbaldwin,tcohn\}@unimelb.edu.au
}

\date{}

\begin{document}
\maketitle

\begin{abstract}
Written text often provides sufficient clues to identify the author, their gender, age, and other important attributes.
Consequently, the authorship of training and evaluation corpora can have unforeseen impacts, including differing model performance for different user groups, as well as privacy implications.
In this paper, we propose an approach to explicitly obscure important author characteristics at training time, such that representations learned are invariant to these attributes.
Evaluating on two tasks, we show that this leads to increased privacy in the learned representations, as well as more robust models to varying evaluation conditions, including out-of-domain corpora.
\end{abstract}

\section{Introduction}
\label{sec:introduction}

Language is highly diverse, and differs according to author, their background, and personal attributes such as gender, age, education and nationality.
This variation can have a substantial effect on NLP models learned from text \cite{hovy2015user}, leading to significant variation in inferences across different types of corpora, such as the author's native language, gender and age.
Training corpora are never truly representative, and therefore models fit to these datasets are biased in the sense that they are much more effective for texts from certain groups of user, e.g., middle-aged white men, and considerably poorer for other parts of the population \cite{hovy2015demographic}.
Moreover, models fit to language corpora often fixate on author attributes which correlate with the target variable, e.g.,  gender correlating with class skews \cite{zhao-EtAl:2017:EMNLP20173}, or translation choices \cite{rabinovich-EtAl:2017:EACLlong}.
This signal, however, is rarely fundamental to the task of modelling language, and is better considered as a confounding influence.
These auxiliary learning signals can mean the models do not adequately capture the core linguistic problem.
In such situations, removing these confounds should give better generalisation, especially for out-of-domain evaluation, a similar motivation to research in domain adaptation based on selection biases over text domains \cite{blitzer2007biographies,daume2007frustratingly}.

%differential-privacy argument.
Another related problem is privacy: texts convey information about their author, often inadvertently, and many individuals may wish to keep this information private.
Consider the case of the \emph{AOL search data leak}, in which AOL released detailed search logs of many of their users in August 2006~\cite{pass2006picture}.
Although they de-identified users in the data, the log itself contained sufficient personally identifiable information that allowed many of these individuals to be identifed~\cite{jones2007know}.
Other sources of user text, such as emails, SMS messages and social media posts, would likely pose similar privacy issues.
This raises the question of how the corpora, or models built thereupon, can be distributed without exposing this sensitive data.
This is the problem of \emph{differential privacy}, which is more typically applied to structured data, and often involves data masking, addition or noise, or other forms of corruption, such that formal bounds can be stated in terms of the likelihood of reconstructing the protected components of the dataset \cite{dwork2008differential}.
This often comes at the cost of an accuracy reduction for models trained on the corrupted data \cite{shokri2015privacy,abadi2016deep}.

Another related setting is where latent representations of the data are shared, rather than the text itself, which might occur when sending data from a phone to the cloud for processing, or trusting a third party with sensitive emails for NLP processing, such as grammar correction or translation.
The transfered representations may still contain sensitive information, however, especially if an adversary has preliminary knowledge of the training model, in which case they can readily reverse engineer the input, for example, by a GAN attack algorithm \cite{hitaj2017deep}. 
This is true even when differential privacy mechanisms have been applied.

Inspired by the above works, and recent successes of adversarial learning \cite{DBLP:conf/nips/GoodfellowPMXWOCB14,ganin2016domain}, we propose a novel approach for privacy-preserving learning of unbiased representations.\footnote{Implementation available at \url{https://github.com/lrank/Robust_and_Privacy_preserving_Text_Representations}.}
Specially, we employ \citeauthor{ganin2016domain}'s approach to training deep models with adversarial learning, to explicitly obscure individuals' private information.
Thereby the learned (hidden) representations of the data can be transferred without compromising the authors' privacy, while still supporting high-quality NLP inference.
We evaluate on the tasks of POS-tagging and sentiment analysis, protecting several demographic attributes --- gender, age, and location --- and show empirically that doing so does not hurt accuracy, but instead can lead to substantial gains, most notably in out-of-domain evaluation.
Compared to differential privacy, we report gains rather than loss in performance, but note that we provide only empirical improvements in privacy, without any formal guarantees.

\section{Methodology}

We consider a standard supervised learning situation, in which inputs 
$\xvec$ are used to compute a representation $\hvec$, which then forms the parameterisation of a generalised linear model,  used to predict the target $\yvec$.
Training proceeds by minimising a differentiable loss, e.g., cross entropy, between predictions and the ground truth, in order to learn an estimate of the model parameters, denoted $\theta_M$.

Overfitting is a common problem, particular in deep learning models with large numbers of parameters, whereby $\hvec$ learns to capture specifics of the training instances which do not generalise to unseen data.
Some types of overfitting are insidious, and cannot be adequately addressed with standard techniques like dropout or regularisation.
Consider, for example, the authorship of each sentence in the training set in a sentiment prediction task.
Knowing the author, and their general disposition, will likely provide strong clues about their sentiment wrt any sentence.
Moreover, given the ease of authorship attribution, a powerful learning model might learn to detect the author from their text, and use this to predict the sentiment, rather than basing the decision on the semantics of each sentence.
This might be the most efficient use of model capacity if there are many sentences by this individual in the training dataset, yet will lead to poor generalisation to test data authored by unseen individuals.

Moreover, this raises privacy issues when $\hvec$ is known by an attacker or malicious adversary.
Traditional privacy-preserving methods, such as added noise or masking, applied to the representation will often incur a cost in terms of a reduction in task performance.
Differential privacy methods are unable to protect the user privacy of $\hvec$ under adversarial attacks, as described in Section \ref{sec:introduction}.

\begin{figure}[t]
    \centering
    \begin{tikzpicture}[line width=0.02cm]

    \node[align=center,minimum size=0.3cm] at (-3.3, 0) (A2) {$\mathbf{x}_i$};

    \node[align=center,minimum size=0.3cm] at (-1.3, 0.25) (B2) {\small $\operatorname{Model} (\theta)$};

    \filldraw[fill=green!20!white, draw=green!20!white,rounded corners] (-0.3, -0.3)rectangle(0.3, 0.3);
    \node[draw=none,align=center,minimum size=0.5cm] at (0.0, +0.0) {\small $\mathbf{h}$};
    \node[draw=none,align=center,text width=6cm] at (1.9,+0.0) {$y_i$};
    \draw[->] (-2.8,  0.0)--( -0.35, -0.0);

    \draw[draw=blue,->] ( 0.35, 0.0)--(1.5, 0.0);
    \node[align=center,minimum size=0.5cm] at (0.8, 0.3) {\small $\theta^c$};

    % Adv

    \draw[draw=red!60!white,dashed,->,rounded corners] (0.1, 0.3)--(0.1, 1.0)--(1.5, 1.0);
    \node[align=center,minimum size=0.3cm] at (0.9, 1.2) {\small $\operatorname{D}_i(\theta^d_i)$};
    \node[align=center,minimum size=0.3cm] at (1.95, 1.0) {$b_i$};

    \draw[draw=red!60!white,dashed,->,rounded corners] (-0.1, 0.3)--(-0.1, 2.0)--(1.5, 2.0);
    \node[align=center,minimum size=0.3cm] at (0.9, 2.2) {\small $\operatorname{D}_j(\theta^d_j)$};
    \node[align=center,minimum size=0.3cm] at (1.95, 2.0) {$b_j$};

\end{tikzpicture}

%%% Local Variables:
%%% mode: latex
%%% TeX-master: "../multidomain_draft"
%%% End:
    \caption{Proposed model architectures, showing a single training instance $(\mathbf{x}_i, y_i)$ with two protected attributes, $b_i$ and $b_j$.
      $\operatorname{D}$ indicates a discriminator, and the \textcolor{red!60!white}{red dashed} and \textcolor{blue}{blue} lines denote adversarial and standard loss, respectively.}
    \label{fig:model}
\end{figure}
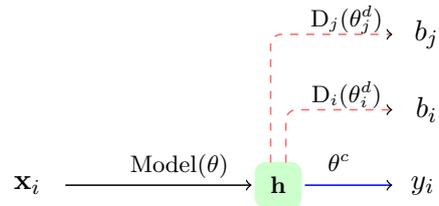

Therefore, we consider how to learn an un-biased representations of the data with respect to specific attributes which we expect to behave as confounds in a generalisation setting.
To do so, we take inspiration from adversarial learning \cite{DBLP:conf/nips/GoodfellowPMXWOCB14,ganin2016domain}.
The architecture is illustrated in \figref{fig:model}.

\subsection{Adversarial Learning}
A key idea of adversarial learning, following \newcite{ganin2016domain}, is to learn a discriminator model $\D$ jointly with  learning the standard supervised model.
Using gender as an example, a discriminator will attempt to predict the gender, $b$, of each instance from $\hvec$, such that training involves joint learning of both the model parameters, $\theta_M$, and the discriminator parameters $\theta_D$.
However, the aim of learning for these components are in opposition -- we seek a $\hvec$ which leads to a good predictor of the target $\yvec$, while being a poor representation for prediction of gender.
This leads to the objective (illustrated for a single training instance),
\begin{align}
  \begin{split}
  \label{equ:sgl}
\hat{\theta} = \min_{\theta_{M}} \max_{\theta_{\D}} ~& \xent( \hat{\yvec}(\xvec; \theta_M), \yvec )  \\
& - \lambda \cdot \xent( \hat{b}(\xvec; \theta_{\D}), b) \, ,
\end{split}
\end{align}
where $\xent$ denotes the cross entropy function.
The negative sign of the second term, referred to as the adversarial loss, can be implemented by a gradient reversal layer during backpropagation \cite{ganin2016domain}.
To elaborate, training is based on standard gradient backpropagation for learning the main task, but for the auxiliary task, we start with standard loss backpropagation, however gradients are reversed in sign when they reach $\hvec$.
Consequently the linear output components are trained to be good predictors, but $\hvec$ is trained to be maximally good for the main task and maximally poor for the auxiliary task. 

Furthermore, \equref{equ:sgl} can be expanded to scenarios with several ($N$) protected attributes,
\begin{align}
 \label{equ:mul}
\hat{\theta} = \min_{\theta_{M}} \max_{ \{ \theta_{\D^i}\}_{i=1}^N} ~& \xent( \hat{\yvec}(\xvec; \theta_M), \yvec )  \\
& - \sum_{i=1}^{N} \left( \lambda_{i} \cdot \xent( \hat{b}(\xvec; \theta_{\D^i}), b_i) \right) \, . \nonumber
\end{align}

\section{Experiments}
In this section, we report experimental results for our methods with two very different language tasks.

\subsection{POS-tagging}
This first task is part-of-speech (POS) tagging, framed as a sequence tagging problem.
Recent demographic studies have found that the author's gender, age and race can influence tagger performance \cite{hovy2015tagging,jorgensen2016learning}.
Therefore, we use the POS tagging to demonstrate that our model is capable of eliminating this type of bias, thereby leading to more robust  models of the problem.

\paragraph{Model}
Our model is a bi-directional LSTM for POS tag prediction \cite{hochreiter1997long}, formulated as:
\begin{align*}
  \hvec_i &  = \operatorname{LSTM} ( \xvec_i, \hvec_{i-1} ; \theta_h) \\
  \hvec'_i &  = \operatorname{LSTM} ( \xvec_i, \hvec'_{i+1} ; \theta'_h) \\
  \yvec_i &  \sim \operatorname{Categorical} (\phi(\left[\hvec_i; \hvec'_i\right]) ; \theta_o) \, ,
\end{align*}
for input sequence $\xvec_i |_{i=1}^n$ with terminal hidden states
$\hvec_0$ and $\hvec'_{n+1}$ set to zero,  where $\phi$ is a linear transformation, and $[\cdot; \cdot]$ denotes vector concatenation.

For the adversarial learning, we use the training objective from
\equref{equ:mul} to protect gender and age, both of which are treated as binary values.
The adversarial component is parameterised by $1$-hidden feedforward nets, applied to the final hidden representation $\left[\hvec_n; \hvec'_0\right]$.
For hyperparameters, we fix the size of the word embeddings and $\hvec$
to $300$, and set all $\lambda$ values to $10^{-3}$.
A dropout rate of $0.5$ is applied to all hidden layers during training.

\paragraph{Data}
We use the \dataset{TrustPilot} English POS tagged  dataset \cite{hovy2015tagging}, which consists of 600 sentences, each labelled with both the sex and age of the author, and manually POS tagged based on the Google Universal POS tagset \cite{petrov2011universal}.
For the purposes of this paper, we follow \citeauthor{hovy2015tagging}'s setup, categorising \resource{sex} into female (\resource{F}) and male (\resource{M}), and \resource{age} into over-45 (\resource{O45}) and under-35 (\resource{U35}).
We train the taggers both with and without the adversarial loss, denoted \resource{adv} and  \resource{baseline}, respectively.

For evaluation, we perform a $10$-fold cross validation, with a train:dev:test split using ratios of 8:1:1.
We also follow the evaluation method in \newcite{hovy2015tagging}, by reporting the tagging accuracy for sentences over different slices of the data based on \resource{sex} and \resource{age}, and the absolute difference between the two settings.

Considering the tiny quantity of text in the \dataset{TrustPilot} corpus, we use the Web English Treebank (\dataset{WebEng}: \newcite{bies2012english}), as a means of pre-training the tagging model.
\dataset{WebEng} was chosen to be as similar as possible in domain to the \dataset{TrustPilot} data, in that the corpus includes unedited user generated internet content. 

As a second evaluation set, we use a corpus of African-American Vernacular English (\dataset{AAVE}) from \newcite{jorgensen2016learning}, which is used purely for held-out evaluation.
\dataset{AAVE} consists of three very heterogeneous domains: \resource{lyrics}, \resource{subtitles} and \resource{tweets}.
Considering the substantial difference between this corpus and \dataset{WebEng} or \dataset{TrustPilot}, and the lack of any domain adaptation, we expect a substantial drop in performance when transferring models, but also expect a larger impact from bias removal using \resource{adv} training.

\begin{table}[t!]
  \sisetup{detect-weight=true,detect-inline-weight=math}
  \centering
  % \footnotesize
  \resizebox{\columnwidth}{!}{%
  \begin{tabular}{
      l
      *{3}{S[table-format=2.1,round-mode=places,round-precision=1]}*{3}{S[table-format=2.1,round-mode=places,round-precision=1]}
    }
    & \multicolumn{3}{c}{\resource{sex}}  & \multicolumn{3}{c}{\resource{age}} \\

\cmidrule(lr){2-4} \cmidrule(lr){5-7}
    & {\resource{F}} & {\resource{M}} & $\Delta$ & {\resource{O45}} & {\resource{U35}} & $\Delta$ \\
\toprule
    \resource{baseline} & 90.949092 & 91.11143 &0.2 & 91.381943    & 89.905378 & 1.5 \\
    \resource{adv}      & \bfseries 92.163674 & \bfseries 92.086631 & \bfseries 0.1 & \bfseries 92.256885  & \bfseries 92.003758 & \bfseries 0.3 \\
    \bottomrule
  \end{tabular}%
  }
  \caption{POS prediction accuracy [\%] using the \dataset{Trustpilot} test set, stratified
    by \resource{sex} and \resource{age} (higher is better), and the
    absolute difference ($\Delta$) within each bias group (smaller is better). The best result is indicated in \textbf{bold}. }
  \label{tab:tag_res}
\end{table}

\paragraph{Results and analysis}
\tabref{tab:tag_res} shows the results for the \dataset{TrustPilot} dataset.
Observe that the disparity for the \resource{baseline} tagger accuracy (the $\Delta$ column), for \resource{age} is larger than for \resource{sex}, consistent with the results of \newcite{hovy2015tagging}.
Our \resource{adv} method leads to a sizeable reduction in the difference in accuracy across both \resource{sex} and \resource{age}, showing our model is capturing the bias signal less and more robust to the tagging task.
Moreover, our method leads to a substantial improvement in accuracy across all the test cases.
We speculate that this is a consequence of the regularising effect of the adversarial loss, leading to a better characterisation of the tagging problem.

\tabref{tab:tag_aave} shows the results for the \dataset{AAVE} held-out domain.
Note that we do not have annotations for \resource{sex} or \resource{age}, and thus we only report the overall accuracy on this dataset.
Note that \resource{adv} also significantly outperforms the \resource{baseline} across the three heldout domains.

Combined, these results demonstrate that our model can learn relatively gender and age \textit{de-}biased representations, while simultaneously improving the predictive performance, both for in-domain and out-of-domain evaluation scenarios.

\subsection{Sentiment Analysis}

The second task we use is sentiment analysis, which also has broad applications to the online community, as well as privacy implications for the authors whose text is used to train our models.
Many user attributes have been shown to be easily detectable from online review data, as used extensively in sentiment analysis results \cite{hovy2015user,potthast2017overview}.
In this paper, we focus on three demographic variables of gender, age, and location.

\paragraph{Model}
Sentiment is framed as a $5$-class text classification problem, which we model using \newcite{kim2014convolutional}'s convolutional neural net (CNN) architecture, in which the hidden representation is generated by a series of convolutional filters followed a maxpooling step, simply denote as $\hvec = \operatorname{CNN} (\xvec ;\theta_M )$.
We follow the hyper-parameter settings of \newcite{kim2014convolutional}, and initialise the model with word2vec embeddings \cite{mikolov2013distributed}.
We set the $\lambda$ values to $10^{-3}$ and apply a dropout rate of $0.5$ to $\hvec$.

As the discriminator, we also use a feed-forward model with one hidden layer, to predict the private attribute(s).
We compare models trained with zero, one, or all three private attributes, denoted \resource{baseline}, \resource{adv}-*, and \resource{adv}-all, respectively.

\begin{table}[t!]
  \sisetup{detect-weight=true,detect-inline-weight=math}
  \centering
  % \footnotesize
  \resizebox{\columnwidth}{!}{%
  \begin{tabular}{
      l
      *{4}{S[table-format=2.1,round-mode=places,round-precision=1]}
    }
    
     & {\resource{lyrics}} & {\resource{subtitles}} & {\resource{tweets}} & {Average} \\
    \toprule
    \resource{baseline} & 73.67483296 & 81.44646656  & 59.86789432 & 71.6630646 \\
    \resource{adv}      & \bfseries 80.46770601 & \bfseries 85.79532026 & \bfseries 65.39231385 & \bfseries 77.02003643   \\
    \bottomrule
  \end{tabular}%
  }
  \caption{POS predictive accuracy [\%] over the \dataset{AAVE} dataset, stratified over the three domains, alongside  the macro-average accuracy. The best result is indicated in \textbf{bold}.}
  \label{tab:tag_aave}
\end{table}

\paragraph{Data}
We again use the \dataset{TrustPilot} dataset derived from \newcite{hovy2015user}, however now we consider the \resource{rating} score as the target variable, not POS-tag.
Each review is associated with three further attributes: gender (\resource{sex}), age (\resource{age}), and location (\resource{loc}).
To ensure that \resource{loc} cannot be trivially predicted based on the script, we discard all non-English reviews based on \resource{langid.py} \cite{lui2012langid},  by retaining only reviews classified as English with a  confidence greater than $0.9$.
We then subsample 10k reviews for each location to balance the five location classes (US, UK, Germany, Denmark, and France), which were  highly skewed in the original dataset. 
We use the same binary representation of \resource{sex} and \resource{age} as the POS task, following the setup in \newcite{hovy2015user}.

To evaluate the different models, we perform $10$-fold cross validation and report test performance in terms of the $F_1$ score for the \resource{rating} task, and the accuracy of each discriminator.
Note that the discriminator can be applied to test data, where it plays the role of an adversarial attacker, by trying to determine the private attributes of users based on their hidden representation.
That is, lower discriminator performance indicates that the representation conveys better privacy for individuals, and vice versa.

\begin{table}[t!]
  \sisetup{detect-weight=true,detect-inline-weight=math}
  \centering
  \resizebox{\columnwidth}{!}{%
  \begin{tabular}{l *{5}{S[table-format=2.1,round-mode=places,round-precision=1]}}
    
           &\multicolumn{2}{c}{$F_1$} & \multicolumn{3}{c}{Discrim. [\%]}\\
    \cmidrule(lr){2-3}     \cmidrule(lr){4-6}
           & {dev} & {test} & {\resource{age}} & {\resource{sex}} & {\resource{loc}} \\
    \toprule
    Majority class                   &       &     & 57.80 & 62.30 & 20.0 \\
    \midrule
    \resource{baseline}            & 41.879 & 40.064  & 65.254 & 66.868 & 53.410 \\
    \midrule
    \resource{adv}-\resource{age}  & \bfseries 42.717 & 40.137  & \bfseries 61.11 & 65.574 & 41.044 \\
    \resource{adv}-\resource{sex}  & 42.351 & 39.903  & 61.756 & 62.893 & 42.708 \\
    \resource{adv}-\resource{loc}  & 42.031 & \bfseries 40.220  & 62.152 & 66.765 & \bfseries 22.097 \\
    \resource{adv}-all             & 42.047 & \bfseries 40.242  & 61.805 & \bfseries 62.468 & 28.149 \\
    \bottomrule
  \end{tabular}%
  }
  \caption{Sentiment $F_1$-score [\%] over the \resource{rating} task, and accuracy [\%] of all the discriminator across three private attributes. The best score is indicated in \textbf{bold}. The majority class with respect to each private attribute is also reported.}
  \label{tab:res_tp}
\end{table}

%%% Local Variables:
%%% mode: latex
%%% TeX-master: "acl_draft"
%%% End:

\paragraph{Results}
\tabref{tab:res_tp} shows the results of the different models.
Note that all the privacy attributes can be easily detected in \resource{baseline}, with results that are substantially higher than the majority class, although \resource{age} and \resource{sex} are less well captured than \resource{loc}.
The \resource{adv} trained models all maintain the task performance of the \resource{baseline} method, however 
they clearly have a substantial effect on the discrimination accuracy.
The privacy of \resource{sex} and \resource{loc} is substantially improved, leading to discriminators with performance close to that of the majority class (conveys little information).
\resource{age} proves harder, although our technique leads to privacy improvements.
Note that \resource{age} appears to be related to the other private attributes, in that privacy is improved when optimising an adversarial loss for the other attributes (\resource{sex} and \resource{loc}).

Overall, these results show that our approach learns hidden representations that hide much of the personal information of users, without affecting the sentiment task performance.
This is a surprising finding, which augurs well for the use of deep learning as a privacy preserving mechanism when handling text corpora.

\section{Conclusion}
We proposed a novel method for removing model biases by explicitly protecting private author attributes as part of model training, which we formulate as deep learning with adversarial learning.
We evaluate our methods with POS tagging and sentiment classification, demonstrating our method results in increased privacy, while also maintaining, or even improving, task performance, through increased model robustness.

\section*{Acknowledgements}
We thank Benjamin Rubinstein and the anonymous reviewers for their helpful feedback and suggestions, and the National Computational Infrastructure Australia for computation resources.
We also thank Dirk Hovy for providing the \dataset{Trustpilot} dataset.
This work was supported by the Australian Research Council (FT130101105).

\bibliographystyle{acl_natbib}
\bibliography{ref}

\end{document}